\def\year{2022}\relax
\documentclass[letterpaper]{article} 
\usepackage{aaai22}  
\usepackage{times}  
\usepackage{helvet}  
\usepackage{courier}  
\usepackage[hyphens]{url}  
\usepackage{graphicx} 
\usepackage{natbib}  
\usepackage{caption} 
\DeclareCaptionStyle{ruled}{labelfont=normalfont,labelsep=colon,strut=off} 
\usepackage{algorithm}
\usepackage{algorithmic}

\usepackage{newfloat}
\usepackage{listings}
\floatstyle{ruled}
\newfloat{listing}{tb}{lst}{}
\floatname{listing}{Listing}
\usepackage{amsmath}
\usepackage{amsthm}
\usepackage{paralist}

\newcommand{\pre}{\textit{pre}}
\newcommand{\tuple}[1]{\ensuremath{\left \langle #1 \right \rangle }}

\usepackage{xspace}

\newcommand{\inoffice}{\textit{in-office}\xspace}
\newcommand{\hasumbrella}{\textit{has-umbrella}\xspace}
\newcommand{\iswet}{\textit{is-wet}\xspace}
\newcommand{\hascoffee}{\textit{has-coffee}\xspace}
\newcommand{\userhascofee}{\textit{user-has-coffee}\xspace}

\newcommand{\buycofee}{\textit{buy-coffee}\xspace}
\newcommand{\movetoofficewithumbrella}{\textit{move-to-office-with-umbrella}\xspace}
\newcommand{\leaveofficewithumbrella}{\textit{leave-office-with-umbrella}\xspace}
\newcommand{\movetoofficewithoutumbrella}{\textit{move-to-office-without-umbrella}\xspace}
\newcommand{\leaveofficewithoutumbrella}{\textit{leave-office-without-umbrella}\xspace}
\newcommand{\getumbrella}{\textit{get-umbrella}\xspace}
\newcommand{\delivercoffee}{\textit{deliver-coffee}\xspace}

\usepackage{booktabs}
\usepackage{multirow}

\title{Learning Probably  Approximately Complete and Safe Action Models for Stochastic Worlds}

\title{An Example of the SAM+ Algorithm for Learning Action Models for Stochastic Worlds}
\author {
    Brendan Juba,\equalcontrib\textsuperscript{\rm 1}
    Roni Stern\equalcontrib\textsuperscript{\rm 2}
}
\affiliations {
    \textsuperscript{\rm 1} Washington University in St.\ Louis\\
    \textsuperscript{\rm 2} Ben Gurion University \& Xerox PARC\\
    bjuba@wustl.edu, rstern@parc.com, sternron@post.bgu.ac.il
}

\usepackage{bibentry}

\begin{document}

\maketitle




In this technical report, we provide a complete example of running the SAM+ algorithm~\cite{juba2022learning}, an algorithm for learning stochastic planning action models, on a simplified PPDDL version of the \emph{Coffee} problem~\cite{dearden1997abstraction}. 
We provide a very brief description of the SAM+ algorithm and detailed description of our simplified version of the Coffee domain. 
For a complete description of SAM+ see Juba and Stern~\shortcite{juba2022learning}.


\subsection{The SAM+ Algorithm}

The SAM+ algorithm takes a set of trajectories $\mathcal{T}$ and a parameter $\delta>0$, and outputs a PPDDL-IP action model denoted $M_\delta$. We describe the preconditions and effects of $M_\delta$ below. 

\noindent \textbf{Preconditions.} 
Let $\mathcal{T}(a)$ be all the action triplets for action $a$. 
States $s$ and $s'$ are said to be a \emph{pre-} and \emph{post-state} of $a$, respectively, if there is an action triplet $\tuple{s,a,s'}\in \mathcal{T}(a)$. 
SAM+ sets the preconditions of an action $a$ to be intersection over all the literals that were true in a pre-state of $a$. 
\begin{equation}
\small
        \pre_{M_\delta}(a) =  \bigcap_{\tuple{s, a, s'}\in \mathcal{T}(a)} s \label{eq:pre} 
\end{equation}        

\noindent \textbf{Effects.} 
Note that we cannot distinguish whether or not $\ell$ was an effect of action $a$ if $\ell\in s$, as it holds in $s'$ in either case. We thus restrict attention to triplets where $\ell\notin s$ to estimate the credal set for $\ell$: 
Let $\#_a(\ell\in s'\setminus s)$ and $\#_a(\ell\notin s)$ 
be the number of action triplets in $\mathcal{T}(a)$ in which $\ell$ is in the post-state and not the pre-state ($|\{\tuple{s, a, s'}\in \mathcal{T}(a):\ell\in s'\setminus s\}|$), 
and the number of action triplets in which $\ell$ was not in the pre-state 
($|\{\tuple{s, a, s'}\in \mathcal{T}(a):\ell\notin s\}|$), respectively. 
SAM+ denotes the intervals $K_{M_\delta}[s'(\ell)|a,s(\ell)]$ 
by $K_\delta(s'(\ell)|a,s(\neg \ell))$, and computes them as follows.
\begin{enumerate}
    \item If $\ell \in \bigcup_{\tuple{s, a, s'}\in \mathcal{T}(a)} s'\setminus s$, 
then 
\begin{equation}
\small
    K_\delta(s'(\ell)|a,s(\neg \ell)) = \frac{\#_a(\ell\in s'\setminus s)}{\#_a(\ell\notin s)} \pm\sqrt{\frac{\ln(2/\delta)}{2\#_a(\ell\notin s)}}
    \label{eq:present-effects}
\end{equation}
\item If $\ell\notin\bigcup_{\tuple{s, a, s'}\in \mathcal{T}(a)} s'\setminus s$, then 
\begin{equation}
\small
    K_\delta(s'(\ell)|a,s(\neg \ell)) = \left[0,\frac{\ln(1/\delta)}{\#_a(\ell\notin s)}\right]
    \label{eq:missing-effects}
\end{equation}
\end{enumerate}
If $\#_a(\ell\notin s)=0$, then $K_\delta(s'(\ell)|a,s(\neg\ell))=[0,1]$. (In any case, we cap the credal sets at $0$ and $1$.) We remark that while the first interval is always valid, the second is smaller and hence preferable for literals we never observe.

Instead of these intervals, it is possible to use the following factors and still maintain reasonable forms of safety: For $\ell\in\bigcup_{\tuple{s, a, s'}\in \mathcal{T}(a)} s'\setminus s $, the transition probability factor for $\ell$ given $\ell\notin s$ and $a$ is an empirical estimate of the probability:
\begin{equation}
\small
\Pr[s'(\ell)|a,s(\neg\ell)]=\frac{|\{\tuple{s, a, s'}\in \mathcal{T}(a):\ell\in s'\setminus s\}|}{|\{\tuple{s, a, s'}\in \mathcal{T}(a):\ell \notin s\}|}
\label{eq:ppddl-exact}
\end{equation}
and otherwise (i.e., for $\ell\notin\bigcup_{\tuple{s, a, s'}\in \mathcal{T}(a)} s'\setminus s $),
\begin{equation}
\small
\Pr[s'(\ell)|a,s(\neg\ell)]=\frac{\ln(2|F||A|/\delta)}{2|\{\tuple{s, a, s'}\in \mathcal{T}(a):\ell\notin s\}|}
\label{eq:ppddl-exact-missing}
\end{equation}
i.e., the midpoints of our previous intervals. 
Using the corresponding PPDDL model enables using PPDDL planners.

\section{The Simplified Coffee Domain}

\begin{figure}
    \centering
\begin{verbatim}
(define (domain simplified-coffee)
(:requirements :negative-preconditions)
(:predicates 
    (in-office) (has-umbrella) (is-wet)
    (has-coffee) (user-has-coffee))

(:action buy-coffee
:precondition (not (in-office))
:effect (and (has-coffee)))

(:action move-to-office-with-umbrella
:precondition (and (not (in-office))
                (has-umbrella))
:effect (and (in-office)))

(:action leave-office-with-umbrella
:precondition (and (in-office)
                (has-umbrella))
:effect (and (not (in-office))))

(:action move-to-office-without-umbrella
:precondition (and (not (in-office))
                (not (has-umbrella)))
:effect (and (in-office)
             (probabilistic 
                0.9 (is-wet))))

(:action leave-office-without-umbrella
:precondition (and (in-office)
                (not (has-umbrella)))
:effect (and (not (in-office))
		     (probabilistic 
		        0.9 (is-wet))))

(:action get-umbrella
:precondition (and (in-office)
                (not (has-umbrella)))
:effect (and (has-umbrella)))

(:action deliver-coffee
:precondition (and (in-office)
                (has-coffee)
                (not (user-has-coffee)))
:effect (and (not (has-coffee) 
            (user-has-coffee)))))
\end{verbatim}
    \caption{PPDDL of the Simplified Coffee domain.}
    \label{fig:coffee-domain}
\end{figure}

The Simplified Coffee domain is simplified version of the Coffee domain introduced by Dearden and Boutilier~\shortcite{dearden1997abstraction}. 
This domain models a robot agent designed to buy coffee from a coffee shop and bring it to the office on a rainy day. 
A state in this domain is defined by five fluents: \inoffice, \hasumbrella, \iswet, \hascoffee, and \userhascofee, abbreviated as IO, HU, IW, HC, and UHC, respectively. 
The agent has 6 actions:
\buycofee, 
\movetoofficewithumbrella, 
\leaveofficewithumbrella, 
\movetoofficewithoutumbrella,
\leaveofficewithoutumbrella, 
\getumbrella,
and 
\delivercoffee, abbreviated as BC, MTOWU, LOWU, MTOWOU, LOWOU, GU, and DC, respectively. 
Figure~\ref{fig:coffee-domain} lists the full PPDDL description of this domain, which includes the preconditions and the stochastic effects of each action. 
For example, \leaveofficewithoutumbrella can only be performed if the agent is in the office and it does not have an umbrella ($IO$ and $\neg HU$), and has the effect of the agent not being in the office ($\neg IO$) and, with probability 0.9, the effect of getting wet ($IW$).

\begin{figure}
    \centering
    \begin{verbatim}
(:define coffee-delivery-problem
    (:init (and 
            (not (user-has-coffee))
            (not (has-coffee))
            (not (has-umbrella))
            (in-office)
            (not (is-wet))
            ))
    (:goal (and (user-has-coffee) 
        (not (is-wet)))))               
    \end{verbatim}
    \caption{PPDDL for a problem in the simplified coffee maker domain.}
    \label{fig:coffee-problem}
\end{figure}
Figure~\ref{fig:coffee-problem} lists the PPDDL description of a problem in the Simplified Coffee domain. 
In this problem, the user does not have coffee yet ($\neg UHC$), 
nor does the agent ($\neg HC$), 
the agent does not have an umbrella ($\neg HU$), it is still in the office ($IO$), 
and it is not wet ($\neg IW$). 
Its goal is to get coffee to the user ($UHC$) without getting wet ($\neg IW$).\footnote{In the original Coffee domain, getting wet introduces a negative reward. Also, note that our goal of avoiding getting wet only states that the agent should not be wet at the goal state. However, since we do not have any action that undoes becoming wet, this means the agent must remain not wet if it aims to achieve this goal.}

\section{SAM+ on the Simplified Coffee Domain}

\begin{table}[]
\begin{tabular}{@{}cccccccc@{}}
\toprule
T                   & State & IO & HU & IW & HC & UHC & Action \\ \midrule
\multirow{2}{*}{T1} & S0    & T  & F  & F  & F  & F   & LOWOU  \\
                    & S1    & F  & F  & T  & F  & F   &   -     \\ \midrule
\multirow{4}{*}{T2} & S0    & T  & F  & F  & F  & F   & LOWOU  \\
                    & S2    & F  & F  & F  & F  & F   & BC     \\
                    & S3    & F  & F  & F  & T  & F   & MTOWOU \\
                    & S4    & T  & F  & T  & T  & F   &   -     \\ \midrule
\multirow{5}{*}{T3} & S0    & T  & F  & F  & F  & F   & LOWOU  \\
                    & S2    & F  & F  & F  & F  & F   & BC     \\
                    & S3    & F  & F  & F  & T  & F   & MTOWU  \\
                    & S5    & T  & F  & F  & T  & F   & DC     \\
                    & S6    & T  & F  & F  & F  & T   &   -     \\ \midrule
\multirow{6}{*}{T4} & S0    & T  & F  & F  & F  & F   & GU     \\
                    & S7    & T  & T  & F  & F  & F   & LOWU   \\
                    & S8    & F  & T  & F  & F  & F   & BC     \\
                    & S9    & F  & T  & F  & T  & F   & MTOWU   \\
                    & S10   & T  & T  & F  & T  & F   & DC     \\
                    & S11   & T  & T  & F  & F  & T   &   -     \\ \bottomrule 
\end{tabular}

\caption{Trajectories T1, T2, T3, and T4 in our example of SAM+ for the Simplified Coffee domain.}\label{tab:trajectories-coffee}
\end{table}

Consider observing the following set of trajectories, all of which start as in the problem listed in Figure~\ref{fig:coffee-problem}, i.e., the agent is in the office without an umbrella, and neither the agent nor the user have coffee. 
In the first three trajectories, denoted T1, T2, and T3, the agent leaves the office without an umbrella. In T1, it gets wet while leaving the office, in T2 it gets wet while moving back to the office after buying the coffee, and in T3 it successfully returns to the office and delivers the coffee. 
Trajectory T4 starts by having the agent picking up an umbrella when it is in the office, and then leaving the office, buying the coffee, returning to the office, and delivering it to the user. Table~\ref{tab:trajectories-coffee} lists these trajectories, including the state before and after each action in them.

Consider the action model learned by applying SAM+ 
after observing each trajectory once, i.e., $\mathcal{T}=\{T1,T2,T3,T4\}$, setting $\delta=0.1$. 
In this model, the preconditions for \leaveofficewithoutumbrella, which has been observed in trajectories T1, T2, and T3, are $\{IO, \neg HU, \neg IW, \neg HC, \neg UHC\}$, matching the initial state S0. 
In general, in SAM+ the effects of each action comprise of literals for which Equation~\ref{eq:present-effects} applies and all other literals, for which  Equation~\ref{eq:missing-effects} applies. 
For the \leaveofficewithoutumbrella action, the literals in the first group are
$\{ \neg IO, IW \}$. 
Applying Equation~\ref{eq:present-effects} on these literals, we get that 
$\neg IO$ is added with probability $1\pm 0.71$ 
and $IW$ is added with probability $1/3\pm 0.71$. 
If we observe each trajectory 100 times, the confidence margins decreases from 0.71 to 0.07. 
The literals $HU$, $HC$, and $UHC$ may also be an effect of \leaveofficewithoutumbrella even though they are not added in any of the trajectories, since they do not appear in the pre-state of \leaveofficewithoutumbrella.   
Applying Equation~\ref{eq:missing-effects} on these literals, we have that in our action model the probability of each of these literals to be added (given that we observe each trajectory 100 times) is in [0,0.008]. 
The corresponding PPDDL action model using Equations~\ref{eq:ppddl-exact} and~\ref{eq:ppddl-exact-missing} yields an action model that assumes 
the effects of \leaveofficewithoutumbrella are to add $\neg IO$ with probability 1.0,
$IW$ with probability 1/3,
$HU$ with probability 0.01,
$HC$ with probability 0.01, 
and $UHC$ with probability 0.01.

The trajectory T1 is more likely than T2 and T3, and so the assumption that we observe the same number of times each trajectory may not be realistic. 
Instead, assume that we observe 895 times T1, 95 times T2, 10 times T3, and 1000 times T4. 
In this case the corresponding effects for \leaveofficewithoutumbrella 
are to add $\neg IO$ with probability 1.0,
$IW$ with probability 0.9,
$HU$ with probability 0.002,
$HC$ with probability 0.002, 
and $UHC$ with probability 0.002. 
Similarly, for the same distribution of observed trajectories (895, 95, 19, and 1000 for T1, T2, T3, and T4, respectively) the preconditions of \movetoofficewithoutumbrella in the learned model are 
$\{\neg IO, \neg HU, \neg IW, HC, \neg UHC\}$ and the 
effects in the corresponding PPDDL domain are 
$IO$ with probability 1.0, 
$IW$ with probabiltiy 0.905, 
and $HU$, $UHC$, and $\neg HC$ each with probability 0.3. 

\section*{Acknowledgements}
We thank our reviewers for their constructive comments.
This research is partially funded by NSF awards IIS-1908287 and CCF-1718380,
and BSF grant \#2018684 to Roni Stern, 
and by the Defense Advanced
Research Projects Agency (DARPA) as part of the SAIL-ON program. 

\bibliography{library}

\end{document}